\documentclass[runningheads]{llncs}
\usepackage[T1]{fontenc}
\usepackage{amsmath, amssymb}
\usepackage{graphicx,verbatim}
\usepackage{hyperref}
\usepackage{cleveref}
\usepackage{glossaries}
\usepackage{multirow, multicol}
\usepackage{booktabs}
\usepackage{enumitem}
\usepackage{delimset}

\usepackage[dvipsnames]{xcolor}

\DeclareMathOperator{\expm}{expm}
\newcommand{\ra}[1]{\renewcommand{\arraystretch}{#1}}
\renewcommand{\phi}{\varphi}

\newcommand{\ie}{\textit{i}.\textit{e}., }

\newcommand{\fb}{\mathbf{f}}

\newcommand{\vb}{\mathbf{v}}
\newcommand{\wb}{\mathbf{w}}
\newcommand{\xb}{\mathbf{x}}
\newcommand{\yb}{\mathbf{y}}
\newcommand{\zb}{\mathbf{z}}

\newcommand{\Ob}{\mathbf{O}}

\newcommand{\Lcal}{\mathcal{L}}

\newcommand{\E}{\mathbb{E}} % Expectation
\newcommand{\Lorentz}{\mathbb{L}} % Lorentz
\newcommand{\R}{\mathbb{R}} % Real numbers
\newcommand{\inner}[2]{\left\langle #1,#2 \right\rangle}

\newcommand{\cbr}[1]{\left\{#1\right\}}
\newcommand{\nbr}[1]{\left\|#1\right\|}

\newacronym{ai}{AI}{Artificial Intelligence}
\newacronym{ml}{ML}{Machine Learning}
\newacronym{dl}{DL}{Deep Learning}
\newacronym{mlr}{MLR}{Multinomial Logistic Regression}
\newacronym{ood}{OOD}{Out-of-Distribution}
\newacronym{resc}{RESC}{Retinal Edema Segmentation Challenge}
\newacronym{lits}{LiTs}{Liver Tumor Segmentation}
\newacronym{ct}{CT}{Computer Tomography}
\newacronym{mri}{MRI}{Magnetic Resonance Imaging}
\newacronym{oct}{OCT}{Optical Coherence Tomography}
\newacronym{knn}{KNN}{K-Nearest Neighbors}

\usepackage{color}

\title{Is Hyperbolic Space All You Need for Medical Anomaly Detection?}

\author{
Alvaro Gonzalez-Jimenez\inst{2,3}%\orcidID{0000-0002-1337-9430}
\and
Simone Lionetti\inst{2}%\orcidID{0000-0001-7305-8957}
\and
Ludovic Amruthalingam\inst{2}%\orcidID{}
\and
Philippe Gottfrois\inst{1,3}%\orcidID{0000-0001-8023-3207}
\and
Fabian Gröger\inst{1,2}%\orcidID{0000-0002-9699-688X}
\and
Marc Pouly\inst{2}\textsuperscript{*}%\orcidID{0000-0002-9520-4799}
\and
Alexander A. Navarini\inst{1,3}\thanks{These authors are joint last authors.}%\orcidID{0000-0001-7059-632X}
}
\authorrunning{Gonzalez-Jimenez et al.}
\institute{
University of Basel, Switzerland \and
Lucerne University of Applied Sciences and Arts, Switzerland \and
University Hospital of Basel, Switzerland \\
\email{alvaro.gonzalezjimenez@usb.ch}
}

\begin{document}

\maketitle

\begin{abstract}
Medical anomaly detection has emerged as a promising solution to challenges in data availability and labeling constraints. Traditional methods extract features from different layers of pre-trained networks in Euclidean space; however, Euclidean representations fail to effectively capture the hierarchical relationships within these features, leading to suboptimal anomaly detection performance. We propose a novel yet simple approach that projects feature representations into hyperbolic space, aggregates them based on confidence levels, and classifies samples as healthy or anomalous. Our experiments demonstrate that hyperbolic space consistently outperforms Euclidean-based frameworks, achieving higher AUROC scores at both image and pixel levels across multiple medical benchmark datasets. Additionally, we show that hyperbolic space exhibits resilience to parameter variations and excels in few-shot scenarios, where healthy images are scarce. These findings underscore the potential of hyperbolic space as a powerful alternative for medical anomaly detection. The project website can be found at \href{https://hyperbolic-anomalies.github.io}{https://hyperbolic-anomalies.github.io}


\end{abstract}
\section{Introduction}\label{sec:introduction}

Anomaly detection and localization plays a critical role in various domains, particularly in medical imaging, where distinguishing and localizing between normal and anomalous samples is crucial. A widely adopted approach involves training models exclusively on healthy images, identifying any deviation from this learned distribution as anomalous \cite{tschuchnigAnomalyDetectionMedical2022}. This strategy mitigates challenges associated with the scarcity of annotated lesion images while reducing annotation costs and biases inherent in training \gls{ai} models.

Among the most effective anomaly detection techniques are projection-based methods, which leverage pre-trained networks to map data into abstract representations, thereby enhancing the separation between normal and anomalous samples. One-class classification \cite{ruffDeepOneClassClassification2018b,liCutPasteSelfSupervisedLearning2021} defines a compact, closed distribution for normal samples, treating any deviations as anomalies. The teacher-student framework \cite{yamadaReconstructionStudentAttention2022,dengAnomalyDetectionReverse2022a} employs a student network to learn normal sample representations from a teacher, using their representation discrepancy to identify anomalies. Memory Bank methods \cite{defardPaDiMPatchDistribution2021a,rothTotalRecallIndustrial2022b,leeCFACoupledhyperspherebasedFeature2022} store normal sample prototypes and apply statistical modeling or distance metrics to detect anomalies.

A common feature across these methods is the extraction of representations from specific layers of a pre-trained network. Each layer encodes hierarchical attributes, but conventional approaches rely on Euclidean space, which may not be the best option to capture hierarchical relationships \cite{nickelPoincareEmbeddingsLearning2017a,salaRepresentationTradeoffsHyperbolic2018a}. This can lead to suboptimal feature representations and reduced anomaly identification performance.

Hyperbolic space, the geometry of constant negative curvature, is well suited to represent hierarchical structures due to its exponential expansion properties \cite{bridsonMetricSpacesNonpositive1999}. Recent advances have demonstrated the effectiveness of hyperbolic embeddings in domains such as few-shot learning \cite{khrulkovHyperbolicImageEmbeddings2020a}, representation learning \cite{desaiHyperbolicImagetextRepresentations2023,ganeaHyperbolicEntailmentCones2018a,lawLorentzianDistanceLearning2019a}, and \gls{ood} detection \cite{vanspenglerPoincareResNet2023a,gonzalez-jimenezHyperbolicMetricLearning2024a}.
Given the hierarchical nature of medical image structures, including disease organization, progression, and anatomical relationships, we hypothesize that hyperbolic space can effectively embed these spatial relationships to enhance anomaly localization. This work aims to answer the following research question: \textbf{Can hyperbolic space effectively represent hierarchical features and improve anomaly localization performance?}

To address this question, we propose a novel framework that generate synthetic anomalies, extracts multi-layer features from a pre-trained network and projects them into hyperbolic space. These hyperbolic embeddings are aggregated by weighting features based on their confidence, specifically considering their distance from the origin \cite{khrulkovHyperbolicImageEmbeddings2020a,ghadimiatighHyperbolicImageSegmentation2022}, which encodes hierarchical depth. Finally, we construct a hyperplane in hyperbolic space to distinguish between normal and anomalous samples.

We validate our framework on multiple medical benchmark datasets including different imaging modalities such as \gls{mri}, \gls{ct}, \gls{oct} and X-Ray. Our results demonstrate that hyperbolic space consistently outperforms Euclidean space for anomaly detection and localization. Additionally, we find that hyperbolic space exhibits robustness to parameter tuning by adaptively learning the optimal curvature, further improving performance. Notably, our approach achieves state-of-the-art results in few-shot settings, where healthy images are scarce or unavailable.

The paper is organized as follows: In \cref{sec:methodology}, we introduce the motivation behind our framework and provide its mathematical formulation. \Cref{sec:experiments} details the datasets, implementation, and training specifics, along with evaluation metrics. \Cref{sec:results} presents experimental findings, comparing hyperbolic and Euclidean-based methods and analyzing performance under few-shot conditions. Finally, \cref{sec:conclusions} summarizes our contributions and discusses broader implications.

\begin{figure}
    \centering
    \includegraphics[width=\textwidth]{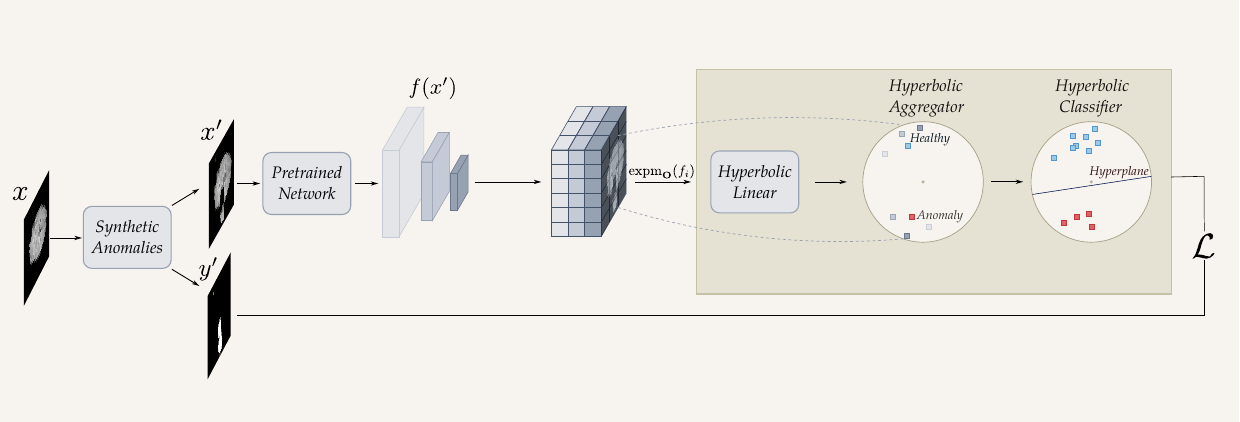}
    \caption{Overview of the anomaly localization methodology in the hyperbolic
    space, from medical anomaly synthesis to classification.}
    \label{fig:main_illustration}
\end{figure}

\section{Methodology}\label{sec:methodology}
This section details our framework to anomaly localization in hyperbolic space, illustrated in \cref{fig:main_illustration}.
In \cref{sec:synthesis-anomalies} we describe our method for synthesizing medical anomalies. 
\Cref{sec:featre-extraction} describes how we obtain patchified features from a pre-trained network.
\Cref{sec:euclidean-to-hyperbolic} outlines the mapping of patchified Euclidean features to hyperbolic space, followed by the hierarchical aggregation process.
Finally, in \cref{sec:hyperbolic-classifier}, we present our hyperbolic classifier, which leverages these aggregated features for classification. 

\subsection{Synthesis Anomalies}\label{sec:synthesis-anomalies}
Given a training set of normal images $x_i \in \mathbb{R}^{H_0 \times W_0 \times C_0}$, we generate images $x'_i$ with synthetic anomalies.
These include
\begin{itemize}[nosep]
    \item \textbf{CutPaste}~\cite{liCutPasteSelfSupervisedLearning2021}, random patches extracted and blended in another location with Poisson image editing~\cite{schluterNaturalSyntheticAnomalies2022,baughManyTasksMake2023};
    \item \textbf{Gaussian Intensity}~\cite{zhangMediCLIPAdaptingCLIP2024}, intensity variations introduced via Gaussian filtering to simulate anomalies such as tumors or cysts;
    \item \textbf{Source Deformation}~\cite{baughManyTasksMake2023}, geometric deformations applied by shifting points within a mask, controlled by a scaling parameter.
\end{itemize}

\subsection{Feature Extraction}\label{sec:featre-extraction}
Features are extracted from the anomaly image $x'_i$ using a pre-trained network, typically a ResNet-like backbone. We select a subset of feature levels $L$, corresponding to different layers in the network hierarchy. At each level, local features are computed by aggregating patch-wise neighborhoods using adaptive average pooling. The resulting feature maps are then upsampled to the highest spatial resolution among them for later aggregation to give a feature map $f_{i,l} \in \R^{C}$.

\subsection{Euclidean to Hyperbolic Features}\label{sec:euclidean-to-hyperbolic}

Hyperbolic geometry, characterized by constant negative curvature, effectively models hierarchical structures~\cite{nickelPoincareEmbeddingsLearning2017a,salaRepresentationTradeoffsHyperbolic2018a}. We employ the Lorentz model due to its simple expression for geodesics~\cite{lawLorentzianDistanceLearning2019a} and numerical robustness~\cite{mishneNumericalStabilityHyperbolic2023}.
Minkowski space is the space of vector $\zb = \brk{z_0, \vec{z}} \in \R\times\R^n$ equipped with the Lorentz inner product $\inner{\zb}{\zb'}_\Lorentz = \vec{z}\cdot\vec{z}' - z_0 z'_0$.

The Lorentz hyperboloid model $\Lorentz_c^n$ of $n$-dimensional hyperbolic space with curvature $c$ is the manifold that satisfies $\inner{\zb}{\zb}_\Lorentz = -1/c$ with $z_0 > 0$.
As the feature vectors lie on the Euclidean space we use the  
the exponential map to projects them onto the hyperboloid
\begin{equation}
    \xb = \expm_\zb(\vb) = \cosh\brk*{\sqrt{c}
    \nbr{\vb}_\Lorentz}\zb +
    \frac{\sinh\brk*{\sqrt{c}\nbr{\vb}_\Lorentz}}{\sqrt{c}\nbr{\vb}_\Lorentz}
    \vb,
\label{eq:exponential-map}
\end{equation}
so $\fb_{i,l} = \expm_\Ob(f_{i,l})$, where $\Ob=\brk{1/\sqrt{c}, \vec{0}}$ is the hyperboloid origin.

% Hyperbolic Aggregation
We project the hyperbolic features to a lower-dimensional hyperbolic space, and adapt the features to the target domain with a hyperbolic linear layer \cite{bdeirFullyHyperbolicConvolutional2023a}, as the network is biased from the pre-training dataset \ie \textit{ImageNet}~\cite{Deng2009} which is suboptimal in medical contexts.
We aggregate features from different hierarchical levels to a single point in hyperbolic space $\zb_i$ using a weighted Lorentzian centroid \cite{lawLorentzianDistanceLearning2019a}:
\begin{equation}
\label{eq:centroid}
 \zb_i =
 \sqrt{c} \frac{\zb'_i}{\abs*{\norm*{\zb'_i}_\Lorentz}}
 \quad\text{with}\quad\zb'_i=\sum_{l\in L} w_{i,l} \fb_{i,l}.
\end{equation}
The weights $w_{i,l}$ are the Euclidean $L_2$ norms of features $\fb_{i,l}$ after transformation to the Poincar\'e ball, which are connected to model confidence \cite{khrulkovHyperbolicImageEmbeddings2020a,ghadimiatighHyperbolicImageSegmentation2022,gonzalez-jimenezHyperbolicMetricLearning2024a}.

% Hyperbolic Classification

\subsection{Hyperbolic Classifier}\label{sec:hyperbolic-classifier}

We classify anomalous features using distances to hyperplanes in the Lorentz model.
The hyperplane in $\Lorentz_c^n$ perpendicular to $\wb$ is given by
\begin{equation}
    H_\wb = \brk[c]{\yb\in\Lorentz_c^n | \inner{\wb}{\yb}_\Lorentz=0},
\end{equation}
and the distance of a point $\zb$ from the hyperplane reads
\begin{equation}
    d_\Lorentz\brk*{\zb, H_\wb} =
    \frac{1}{\sqrt{c}} \abs*{ \sinh^{-1} \brk*{\sqrt{c} \frac{\inner{\wb}{\zb}_\Lorentz}{\norm{\wb}_\Lorentz}}}.
\end{equation}
The patch-wise logit and probability for an image $x_i$ with representation $\zb_i$ to be anomalous are then given by
\begin{equation}
    \ell_{\wb}(\zb_i) = \text{sign}\brk*{\inner{\wb}{\zb_i}_\Lorentz} \norm{\wb}_\Lorentz d_\Lorentz\brk*{\zb_i, H_{\wb}},
    \qquad
    p_{\wb}(\zb_i) = \brk[s]*{1 + \exp\brk{\ell_{\wb}(\zb_i)}}^{-1}.
\end{equation}
The model constructs a hyperplane for robust class discrimination by optimizing it through binary cross-entropy
\begin{equation}
 \Lcal
 = - \E_{x_i \sim \mathcal{A}} \brk[s]*{\log\brk{p_{\wb}(\zb_i)}}
 - \E_{x_i \sim \mathcal{N}} \brk[s]*{\log\brk{1 - p_{\wb}(\zb_i)}}   
\end{equation}
where $\mathcal{A}$ and $\mathcal{N}$ are sets of anomalous and normal pixel centroids computed via \cref{eq:centroid}, respectively.
\section{Experiments}
\label{sec:experiments}

This section describes how hyperbolic space is evaluated for anomaly detection and localization.

\subsection{Datasets}

We follow BMAD~\cite{baoBMADBenchmarksMedical2024}, a recent benchmark for medical anomaly detection and localization spanning different imaging modalities.
It features defined dataset splits to facilitate reproducibility and prevent leakage.
We only deviate by excluding the pathology dataset Camelyon16 due to known difficulties with memory, and by resizing all images to $224\times224$ pixels.
The five datasets used in this work are summarized in \cref{tab:datasets}.

\textbf{BraTS2021}~\cite{baidRSNAASNRMICCAIBraTS20212021} is a widely used dataset for brain tumor segmentation and classification in \gls{mri}, BMAD considers the FLAIR sequences for anomaly detection.
% Brats2021/valid/good=39
% Brats2021/valid/anomaly=44
% Brats2021/test/good=640
% Brats2021/test/anomaly=3075

\textbf{BTCV}~\cite{synapse} and \textbf{\gls{lits}}~\cite{bilicLiverTumorSegmentation2023} focus on liver \gls{ct} imaging. BMAD uses the anomaly-free BTCV set for training and LiTs for evaluation.
% Liver/valid/good=93
% Liver/valid/anomaly=73
% Liver/test/good=833
% Liver/test/anomaly=660
 
The \textbf{\gls{resc}}~\cite{huAutomatedSegmentationMacular2019} provides \gls{oct} images for retinal pathology analysis.
% RESC/valid/good=45
% RESC/valid/anomaly=70
% RESC/test/good=1041
% RESC/test/anomaly=764

\textbf{OCT2017}~\cite{kermanyIdentifyingMedicalDiagnoses2018} is a large-scale OCT dataset for retinal disease classification, comprising one normal category and three medical conditions. The latter are treated as a single abnormal class.
% OCT2017/valid/good=8
% OCT2017/valid/anomaly=24
% OCT2017/test/good=242
% OCT2017/test/anomaly=726

\textbf{RSNA}~\cite{wangChestXRay8HospitalScaleChest2017} contains chest X-rays labeled with one normal category and eight conditions, all of which are treated as a single abnormal class.
% RSNA/valid/good=70
% RSNA/valid/anomaly=1420
% RSNA/test/good=781
% RSNA/test/anomaly=16413

\begin{table}
	\caption{Count of normal and anomalous samples across BMAD dataset splits.}
        \label{tab:datasets}
	\centering
	\ra{1.3}
	\resizebox{\textwidth}{!}{%
	   \begin{tabular}{lrrrrrrrrrrrrrrr}
            \toprule
			Dataset & \phantom{a} &
			\multicolumn{2}{c}{BraTS2021} &
			\phantom{a} &
			\multicolumn{2}{c}{BTCV$+$LiTs} &
			\phantom{a} &
			\multicolumn{2}{c}{RESC} &
			\phantom{a} &
			\multicolumn{2}{c}{OCT2017} &
			\phantom{a} &
			\multicolumn{2}{c}{RSNA} \\
			\cmidrule{3-4}
			\cmidrule{6-7}
			\cmidrule{9-10}
			\cmidrule{12-13}
			\cmidrule{15-16}
            Split $\downarrow$ & & norm. & anom. & & norm. & anom. & & norm. & anom. & & norm. & anom. & & norm. & anom. \\ \midrule
            Train & & \phantom{0}7,500 & \phantom{00,00}0 & & \phantom{0}1,542 & \phantom{00,00}0 & & \phantom{0}4,297 & \phantom{00,00}0 & & 26,315 & \phantom{00,00}0 & & \phantom{0}8,000 & \phantom{00,00}0\\
            Valid & & 39 & 44 & & 93 & 73 & & 45 & 70 & & 8 & 24 & & 70 & 1,420\\
            Test & & 640 & 3,075 & & 833 & 660 & & 1,041 & 764 & & 242 & 726 & & 781 & 16,413\\
		\bottomrule
		\end{tabular}
	}
\end{table}

\subsection{Experimental Setup}

We use a pre-trained WideResNet50~\cite{zagoruyko2016wide} as feature extractor in all experiments. To ensure a fair comparison with baseline methods, we refrain from using data augmentation, applying only ImageNet-based normalization \cite{Deng2009}. The pre-trained network is frozen, and only the hyperbolic components are trained. We extract features from \texttt{layer\_2} and \texttt{layer\_3}, with a dimensionality of 1024, which are subsequently patchified using a patch size of 3.

The curvature parameter is trainable and initialized to $c=1$. Training is conducted for 50 epochs across all datasets using the Adam optimizer with a learning rate of $10^{-3}$ and a batch size of 32. All experiments are performed on a single NVIDIA Tesla V100 GPU with 32 GB of memory.

We evaluate both image-level (detection) and pixel-level (localization) performance using Image-AUROC ($\text{I}_{\text{AUROC}}$) and Pixel-AUROC ($\text{P}_{\text{AUROC}}$) in percentage, respectively. In addition to our hyperbolic approach, we benchmark against several state-of-the-art Euclidean anomaly detection and localization models, including RD4AD~\cite{dengAnomalyDetectionReverse2022a}, STFPM~\cite{yamadaReconstructionStudentAttention2022}, PaDiM~\cite{defardPaDiMPatchDistribution2021a}, PatchCore~\cite{rothTotalRecallIndustrial2022b}, and CFA~\cite{leeCFACoupledhyperspherebasedFeature2022}.

Finally, we test for statistical significance using the Mann-Whitney U test to compare the AUROC distributions between two models. We assume statistical significance for $p < 0.05$ and denote this with \textbf{bold}.

% Results section
\section{Results}\label{sec:results}

\begin{table}
	\caption{Comparison of anomaly detection and localization performance across medical datasets.
The values represent the mean, the minimum (subscript), and maximum (superscript) over 5 different random seeds.}
        \label{tab:results_anomaly}
	\centering
	\ra{1.3}
	\resizebox{\textwidth}{!}{%
	   \begin{tabular}{lcccccccccccc}
            \toprule
			\multirow{2}{*}{\parbox[c]{.15\linewidth}{\centering Methods}} &
			\multicolumn{2}{c}{BraTS2021}                                  &
			\phantom{abc}                                                  &
			\multicolumn{2}{c}{BTCV $+$ LiTs}                              &
			\phantom{abc}                                                  &
			\multicolumn{2}{c}{RESC}                                       &
			\phantom{abc}                                                  &
			\multicolumn{1}{c}{OCT2017}                                    &
			\phantom{abc}                                                  &
			\multicolumn{1}{c}{RSNA}                                       \\
			\cmidrule{2-3}
			\cmidrule{5-6}
			\cmidrule{8-9}
			\cmidrule{11-11}
			\cmidrule{13-13}
                & $\text{I}_{\text{AUROC}}$ & $\text{P}_{\text{AUROC}}$ &   &
                $\text{I}_{\text{AUROC}}$                                      & $\text{P}_{\text{AUROC}}$ &                           &
                $\text{I}_{\text{AUROC}}$                                      & $\text{P}_{\text{AUROC}}$ &                           &
                $\text{I}_{\text{AUROC}}$                                      &                           &
                $\text{I}_{\text{AUROC}}$ \\ \midrule
    RD4AD                                                          & $89.52_{88.85}^{90.19}$         & $96.36_{96.24}^{96.48}$         &   & $59.14_{53.83}^{64.45}$ & $91.40_{91.30}^{91.50}$  &  & $88.25_{86.25}^{90.25}$ & $96.18_{95.98}^{96.38}$ &  & $94.88_{92.17}^{97.58}$ &  & $67.63_{66.53}^{68.73}$ \\
    STFPM                                                          & $84.25_{81.87}^{86.63}$         & $96.03^{96.43}_{95.63}$         &   & $61.48_{59.81}^{63.15}$ & $96.26_{96.12}^{96.40}$  &  & $87.26_{87.03}^{87.49}$ & $94.96_{94.90}^{95.02}$ &  & $91.88_{90.55}^{93.21}$ &  & $69.31_{68.22}^{70.4}$ \\
    PaDiM                                                          & $79.62_{78.28}^{80.96}$         & $94.22_{93.99}^{94.45}$         &   & $50.91_{50.58}^{51.24}$ & $90.48_{90.33}^{90.63}$  &  & $75.15_{73.73}^{76.57}$ & $91.22_{90.85}^{91.59}$ &  & $90.17_{89.56}^{90.78}$ &  & $74.48_{74.22}^{74.74}$ \\
    PatchCore                                                      & $92.02_{91.91}^{92.13}$                         & $95.53_{95.48}^{95.58}$                         &   & $59.33_{59.19}^{59.47}$                 & $95.00^{95.01}_{94.99}$                  &  & $90.54_{90.44}^{90.64}$                 & $95.87_{95.83}^{95.91}$                 &  & $97.45_{96.80}^{98.10}$                 &  & $75.67_{75.47}^{75.87}$                 \\
    CFA                                                            & $84.99_{84.83}^{85.15}$         & $96.61_{96.57}^{96.65}$         &   & $53.89_{49.65}^{58.13}$ & $97.40_{97.34}^{97.46}$  &  & $72.47_{70.20}^{74.74}$ & $92.49_{91.41}^{93.57}$ &  & $79.10_{78.54}^{79.66}$ &  & $66.65_{66.50}^{66.80}$ \\
    Ours                                                           & $92.49_{91.96}^{93.02}$         & $95.56_{95.49}^{95.63}$         &   & $\mathbf{65.94_{63.89}^{67.99}}$ & $96.49_{93.87}^{99.11}$  &  & $90.71_{90.14}^{91.28}$ & $95.32_{95.08}^{95.56}$ &  & $97.85_{97.58}^{98.12}$ &  & $\mathbf{79.46_{78.72}^{80.20}}$ \\
		\bottomrule
		\end{tabular}
	}
\end{table}

\Cref{tab:results_anomaly} presents the experimental results for anomaly detection and localization.
We observe acceptable agreement of the methods based on Euclidean geometry with the results reported by the BMAD in table 2 \cite{baoBMADBenchmarksMedical2024}.
Among these, PatchCore achieves the most consistent performance across datasets, although the somewhat lower performance for BTCV$+$\gls{lits} could be interpreted as sensitivity to a distribution shift.
In contrast, the proposed hyperbolic framework shows the best performance on whole images across all datasets, even if this is not always statistically significant, and it remains robust even for BTCV$+$\gls{lits}.
At the pixel level, the hyperbolic approach remains competitive with other methods, even though different Euclidean baselines outperform it in specific cases. However, in medical practice, misdiagnosing an entire image is generally more problematic than minor pixel-wise mismatches.

\subsection{Ablation Study on Model Parameters}
\begin{figure}
	\centering
	\includegraphics[width=\textwidth]{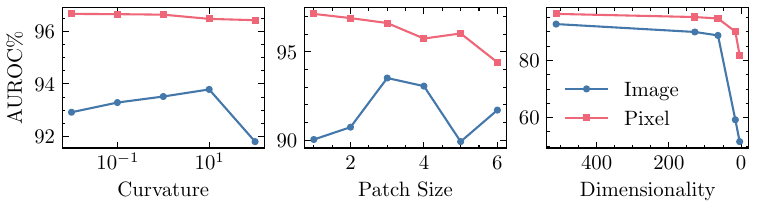}
    \caption{Ablation study on key components of our hyperbolic framework: fixed curvature, patch size variations, and hyperbolic layer dimensionality.}
	\label{fig:ablation_studies}
\end{figure}

We conduct ablation studies on curvature, patch size, and dimensionality of the hyperbolic space using the BraTS dataset. \Cref{fig:ablation_studies} presents the impact of these variations on performance.

%Curvature Adaptation
We first investigate the role of curvature by fixing it to $c=\brk[c]{0.01,0.1,1,10,100}$.
The first plot indicates that constraining the curvature leads to a decline in performance, with better results observed at lower curvature values. This underscores the advantage of a learnable curvature, which allows the model to adaptively optimize the geometry of the representation space for anomaly identification.

%Patch Size Variation:
Next, we analyze the effect of the patch size 
$\brk[c]{1,2,3,4,5,6}$ when aggregating local features. Increasing the patch size negatively impacts both $I_\text{AUROC}$ and $P_\text{AUROC}$. This suggests that fine-grained feature extraction is preferable for capturing subtle anomalies, whereas overly large patches may dilute local information critical for accurate anomaly localization.

%Hyperbolic reduction
Lastly, hyperbolic space has been shown to efficiently encode representations in lower-dimensional embeddings, making it advantageous for memory-constrained scenarios \cite{kusupatiMatryoshkaRepresentationLearning2024,gonzalez-jimenezHyperbolicMetricLearning2024a}. To evaluate this, we reduce the feature dimensionality to 
$\brk[c]{512,128,16,8,2}$. The last plot reveals that while $I_\text{AUROC}$ is more sensitive to extreme dimensionality reduction, $P_\text{AUROC}$ remains relatively stable.

\subsection{Few-Shot Anomaly Detection and Localization}
\begin{figure}
	\centering
	\includegraphics[width=.9\textwidth]{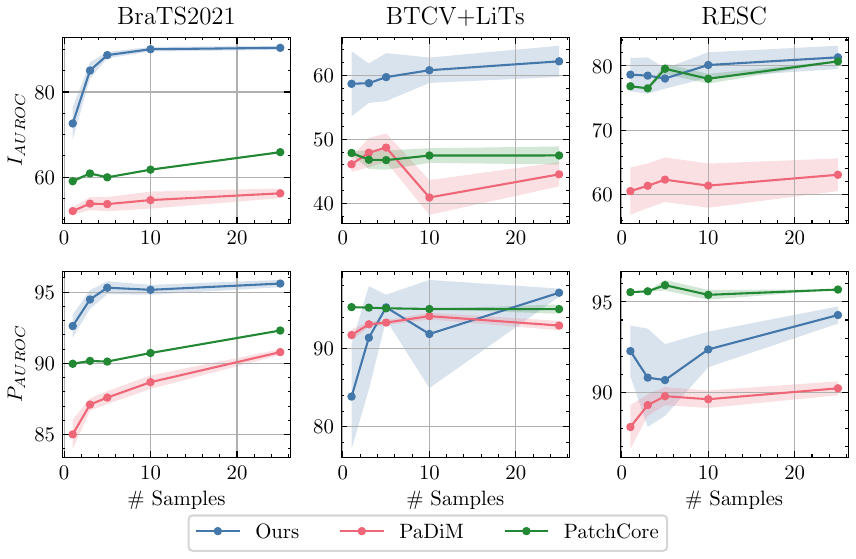}
	\caption{Few-shot evaluation with varying normal image counts $\brk[c]{1,3,5,10,25}$. Our hyperbolic model outperforms PaDiM and PatchCore in scarce data scenarios.
    Error bands are obtained with five different random seeds, without changing the training set.}
	\label{fig:few_shot}
\end{figure}

We further evaluate the robustness of our framework in a few-shot setting, where only a limited number of normal images are available for training. We experiment with $\cbr{1,3,5,10,25}$ normal images and compare our performance against PaDiM~\cite{defardPaDiMPatchDistribution2021a} and PatchCore~\cite{rothTotalRecallIndustrial2022b}. The results in \Cref{fig:few_shot} demonstrate that our hyperbolic model significantly outperforms both baselines, particularly in extreme data scarcity scenarios.

% Conclusions
\section{Conclusions}\label{sec:conclusions}

In this work, we introduced a hyperbolic anomaly detection and localization framework that leverages the unique geometric properties of hyperbolic space to enhance both classification and localization of medical anomalies. 

Our evaluation across multiple medical imaging datasets demonstrates that our method consistently outperforms state-of-the-art anomaly identification approaches in terms of $I_\text{AUROC}$, and is competitive with the best ones for localization as demonstrated by $P_\text{AUROC}$. Additionally, we show that hyperbolic embeddings retain strong performance in low dimensions enabling efficient deployment in resource-constrained environments, and consistently outperforms Euclidean baselines in few-shot data regimes.

% Limitations of the work
One key area for future investigation is the incorporation of features from earlier layers of the model, which could help leverage the hierarchical information embedded throughout the network. Although fully hyperbolic networks have been shown to outperform hybrid architectures \cite{bdeirFullyHyperbolicConvolutional2023a,chen-etal-2022-fully,vanspenglerPoincareResNet2023a}, this remains a developing research area, and challenges related to stability and reproducibility persist.
% Metrics limitations
Furthermore, while our study focused on feature-based anomaly baselines, expanding the comparison to reconstruction-based \cite{livernoche2023diffusion,liu2025survey} or gradient-based \cite{gonzalezjimenezSANO} with other performance metrics could provide a more comprehensive assessment of the model's effectiveness.

% Multimodality
Additionally, integrating multi-modal data, such as radiology reports or genomic information, could provide a richer anomaly characterization, improving interpretability and clinical utility.

Our work contributes to the advancement of AI-driven medical anomaly detection and localization, with a particular emphasis on improving the accuracy and localization of anomalies, especially in few-shot settings. These findings have the potential to significantly enhance medical image quality assessment and facilitate the quantitative analysis of rare diseases, ultimately leading to more precise and data-efficient diagnostic models for clinical applications.

\begin{credits}
.
\subsubsection{\discintname}
The authors have no competing interests to declare that are
relevant to the content of this article.

\end{credits}
\bibliographystyle{splncs04}
\bibliography{bibliography}

\end{document}